# Planning in POMDPs Using Multiplicity Automata

**Eyal Even-Dar**
School of Computer Science
Tel-Aviv University
Tel-Aviv, Israel 69978
evend@post.tau.ac.il

**Sham M. Kakade**
Computer and Information Science
University of Pennsylvania
Philadelphia, PA 19104
skakade@linc.cis.upenn.edu

**Yishay Mansour**
School Computer Science
Tel-Aviv University
Tel-Aviv, Israel 69978
mansour@post.tau.ac.il

## Abstract

Planning and learning in Partially Observable MDPs (POMDPs) are among the most challenging tasks in both the AI and Operation Research communities. Although solutions to these problems are intractable in general, there might be special cases, such as structured POMDPs, which can be solved efficiently. A natural and possibly efficient way to represent a POMDP is through the predictive state representation (PSR) — a representation which recently has been receiving increasing attention.

In this work, we relate POMDPs to multiplicity automata — showing that

POMDPs can be represented by multiplicity automata with no increase in the representation size. Furthermore, we show that the size of the multiplicity automaton is equal to the rank of the predictive state representation. Therefore, we relate both the predictive state representation and POMDPs to the well-founded multiplicity automata literature.

Based on the multiplicity automata representation, we provide a planning algorithm which is exponential only in the multiplicity automata rank rather than the number of states of the POMDP. As a result, whenever the predictive state representation is logarithmic in the standard POMDP representation, our planning algorithm is efficient.

## 1 Introduction

In the last decade, the field of reinforcement learning has become a dominant tool for formulating and solving many real world sequential decision problems. The most popular mathematical framework for formulating this problem is the Markov Decision Process (MDP). Here, planning can be done efficiently using algorithms such as policy iteration, value iteration and linear programming, and, when the state space is so large that these algorithms become intractable, there are a number of other approximate techniques (e.g. Sutton and Barto [1998], Bertsekas and Tsitsiklis [1996]).

Although the MDP framework is appealing, it carries the often unrealistic assumption that the system state is fully observable by the agent. Unfortunately, in many cases where the system state is not revealed to the agent, the MDP framework is not applicable and a different model is in order. In such cases, the Partially Observable Markov Decision Process (POMDP) provides a suitable mathematical framework.

Unlike for MDPs, the problem of planning in POMDPs is intractable in general. For example, it is NP-Hard to give an algorithm, which computes an $\epsilon$-optimal policy in a POMDP Lusena et al. [2001]. Although the problem is intractable, many heuristics and approximation algorithm have been proposed. One such class of approximation schemes are grid based algorithms for POMDPs (See Lovejoy [1991], Brafman [1997], Hauskrecht [1997], Bonet [2002] and references therein). In essence, these grid based algorithms discretize the *belief states* in the POMDP in some manner into a finite state MDP. Recall that in a POMDP, the belief state — the distribution over the states given the observable — is what is relevant and there are an (uncountably) infinite number of such states.

Recently, an alternative and more concise representation for POMDPs, namely *Predictive State Representations* (PSRs) (Littman et al. [2001], Singh et al. [2003, 2004]), has been proposed. The hope is that this representation will allow for more efficient approximation algorithms. In the PSR, the POMDP, "hidden state" structure is not explicitly represented and only the probabilities of future, "action - observation" sequences are specified.

In this work, we focus on the closely related representation of multiplicity automata, which were first introduced by Shlitzenberger [1961] and were closely studied by Fliess

[1974], Carlyle and Paz [1971]. Multiplicity automata are generalizations of both deterministic and stochastic automata, and we show that they are also generalizations of the POMDPs.

Our first contribution is in showing that POMDPs are a special case of multiplicity automata. This is done through using a similar construction to the PSR construction. Similar to PSRs, we then show that a POMDP can be represented by multiplicity automaton of size $r$, where $r$ is bounded by the number of the POMDP states. These results formally relate PSRs to a large extant literature on that of multiplicity automata, where many learning models have been considered (for instance, see Beimel et al. [2000]).

Our second contribution is to provide a planning algorithm which exploits this structure. In particular, we provide an algorithm which has runtime that is exponential in the multiplicity automata size rather than the number of states of the POMDP. This result directly contributes to the extant PSR literature — essentially, in the PSR framework, our algorithm can be viewed as providing an approximate planning algorithm with runtime that is exponential in the PSR rank, which, to our knowledge, is a novel result in this literature.

Our algorithm differs from other grid based ones in that it uses a *modified belief MDP* and not the standard *belief MDP*. The *modified belief MDP* is built with respect to the multiplicity automaton construction. Since previous planning based algorithms have runtime which is exponential in the number of states of the POMDP, our algorithms provides a significant improvement whenever the multiplicity automata representation is logarithmic in the standard POMDP representation. In such cases, the planning problem becomes tractable. An example for POMDP that has a logarithmic PSR representation was given by Littman et al. [2001] (the example is the POMDP extension of the factored MDP that was introduced by Singh and Cohn [1998]).

We now briefly outline the difficulty in our approach, as opposed to standard grid based approaches. Consider a standard belief state representation in which a belief state is represented as a vector, which specifies the probability distribution over the $n$ states (so the vector is in $\mathbb{R}^n$ and sums to 1). It is easy to see that there exists an orthonormal basis for the belief states, with the property that every belief state can be represented as a linear combination of these basis vectors with weights that are bounded (by 1). In the multiplicity automata (and also the PSR) representation, belief states are also represented as vectors, except now the vectors are not probability vectors (they may have negative entries and not to sum to one). Here, one can show that these vectors are spanned by a basis with size that is no greater than the number of states. However, the problem is that representing any state as linear combination of these basis vectors might involve using arbitrarily large weights, which makes a discretization scheme problematic. Here, we show to how to construct a basis which spans the set of all states with "small" weights (using the methods in Awerbuch and Kleinberg [2004]). Using this carefully constructed basis set, we devise our planning algorithm.

The paper is organized as follows. Section 2 provides a formal definition of POMDPs. Grid based algorithms for POMDPs are defined and analyzed in Subsection 2.2. Then, in Section 3, we provide a definition and concrete examples of multiplicity automate. The inherent connection between multiplicity automata and POMDPs is explored in Section 4. In section 5 we show how given a POMDP we can build its matrix representation compactly and in Subsection 5.3 we show how to improve this representation. Finally, in Section 6, we provide an algorithm which is exponential only on the multiplicity automaton rank.

## 2 Model

### 2.1 Partially Observable Markov Decision Processes

A *Partially Observable Markov Decision Process* (POMDP) consists of:

- A finite set of states $S = \{s_1, \ldots, s_n\}, n = |S|$.

- A finite set of actions $A(s_i)$ associated with each state $s_i$. We assume identical action sets, i.e., for every state $s_i$, $A(s_i) = A$

- The transition probability matrix, $P(s, a, s')$, which is the probability of moving to state $s'$ after performing action $a$ at state $s$.

- A finite set of observable signals $O \times R$ where $O$ is the observation set and $R$ is the set of immediate rewards. We assume each immediate reward is bounded in $[0, 1]$. For simplicity, we assume that the possible observation signals are identical for all states.

- An observation probability, given a state $s_i$ and action $a \in A(s_i)$, for an observation $o \in O \times R$, is $OB(o|s_i, a)$.

We define the return of a policy $\pi$ in a POMDP, $R^\pi$ as the expected sum of its discounted rewards, where there is a discount factor $\gamma \in (0, 1)$, i.e., $R^\pi(s) = E[\sum_{t=0}^\infty \gamma^t r_t | \pi, s]$.

An equivalent problem to finding the optimal strategy in a POMDP is solving the associated *Belief MDP*. The Belief MDP is a continuous state MDP, in which the states are distributions over the original POMDP states, i.e., a state $x$ is a distribution over $S$ such that $x(i)$ is the probability of being at state $s_i$. These distributions over states are referred to as *belief states*. The transition probabilities in this

MDP are defined according to the MDP transition and observation probability (The update is bayesian). Similarly to finite state MDP we define the value function under policy $\pi$ from state $s$ as $V^\pi(s) = E[\sum_{t=0}^\infty \gamma^t r_t | \pi, s]$ and the Q-function as $Q^\pi(s,a) = E[\sum_{t=0}^\infty \gamma^t r_t | \pi, s, a_0 = a]$. We denote the optimal policy as $\pi^*$ and denotes its value and Q function as $V^*$ and $Q^*$.

## 2.2 Grid Based Algorithms for POMDPs

Grid based algorithms for POMDPs are a standard and useful technique in solving POMDPs. The grid based algorithms discretize the belief states in the Belief MDP into a finite set, and they differ by the way they discretize it. The discreteization is done by defining a mapping from the *belief states* to the grid states, i.e., $g : B \to G$, where $G$ is the grid states. The grid mesh parameter is defined as $\delta = \sup_{b, b', g(b) = g(b')} \|b - b'\|_1$. Unfortunately, all grid based algorithms suffer from an exponential dependence on the number of states that is the grid is composed form $(\frac{1}{\delta})^n$ cells where $\delta$ is the grid mesh.

To give intuition we provide here a simple grid based algorithm and performance guarantees for it. Similar results can be found in the literature. First we discretize the *belief space* by having $(\frac{1}{\delta})^n$ grid states. Each state is a vector of the form $(\alpha_1, ..., \alpha_n)$, where $\alpha_i = m\delta$, $m \in \{0, 1, \ldots, 1/\delta\}$.

The MDP is constructed from these discretized states in the natural way. The transition probabilities are now just modified such that transition only occur between discretized belief states. We call the resulting MDP the $\delta$-discretized MDP of the POMDP. Under this modification, it is relatively straightforward to show the following theorem.

**Theorem 2.1** *Let $M$ be the $\delta$-discretized MDP for a POMDP. Then the optimal policy in $M$ is $\frac{2\delta}{(1-\gamma)^3}$-optimal policy in the POMDP.*

The following three simple technical lemma establish the proof.

**Lemma 2.2** *Let $x$ and $y$ be any two belief states such that $\|x - y\|_1 \leq \delta$, then $|Q^*(x,a) - Q^*(y,a)| \leq \frac{\delta}{1-\gamma}$*

**Proof:** Consider any policy $\pi$ and start it from state $x$ and from state $y$. Since $\|x - y\|_1 \leq \delta$ the expected reward at time 1 differs by at most $\delta$. Let $x_t, y_t$ be the belief states at time $t$ after starting at states $x, y$ respectively and following policy $\pi$ for $t$ steps then $\|x_t - y_t\|_1 \leq \delta$. The proof is done by induction. The basis holds since $x_1 = x$ and $y_1 = y$. Assume the hypothesis holds for $t-1$ and prove for $t$, since $\|x_{t-1} - y_{t-1}\|_1 \leq \delta$, then after following $\pi$ for one step w.p $1 - \delta$ for both states we will have the same behavior, thus $\|x_t - y_t\|_1 \leq \delta$ and this implies that the reward difference is bounded by $\delta$ as well at time $t$. □

This leads to a standard results on the value attained by value iteration on aggregated states (see Bertsekas and Tsitsiklis [1996] page 350).

**Lemma 2.3** *Let the grid $G$ defined by a function $g : B \to G$, and suppose that for every $b, b'$ such that $g(b) = g(b')$, $\|b - b'\|_1 \leq \delta$. If we take one state as representative in each cell and let $\tilde{V}^*$ be the optimal value in this MDP then we have*

$$|V^*(b) - \tilde{V}^*(g(b))| \leq \frac{\delta}{(1-\gamma)^2}$$

The next lemma is also standard and relates the greedy policy loss to the value function approximation error (see Singh and Yee [1994], Bertsekas and Tsitsiklis [1996]).

**Lemma 2.4** *If $\tilde{Q}$ is a function such that $|\tilde{Q}(s,a) - Q^*(s,a)| \leq \delta$ for all $s \in S$ and $a \in A$. Then for all $s$*

$$V^*(s) - V^{\tilde{\pi}}(s) \leq \frac{2\delta}{1-\gamma},$$

*where $\tilde{\pi} = Greedy(\tilde{Q})$.*

Now the proof of Theorem 2.1 follows, by Lemma 2.2 if the grid mesh is $\delta$ then the difference in $Q^*$ in the same grid state is at most $\delta/(1-\gamma)$ and then we can simply apply Lemmas 2.3 and 2.4.

## 3 Multiplicity Automata

In this section, we define the multiplicity automata and provide examples of other automata which are special case of the multiplicity automata.

Starting with notation, let $K$ be a field[1] (in this paper, we only use $K = \mathbb{R}$), $\Sigma$ an alphabet, $\lambda$ the empty string, and $f : \Sigma^* \to K$ be a function (here, $\Sigma^*$ is the set of all strings comprised of elements from $\Sigma$). Let us now define the Hankel matrix, which corresponds to $f$ (sometimes referred to as a formal series).

**Definition 3.1** *Associate with $f$ an infinite Hankel matrix $F$, where each row is indexed by some $x \in \Sigma^*$ and each column is indexed by some $y \in \Sigma^*$. Let $F(x,y) = f(x \circ y)$, where $\circ$ denotes the concatenation operation.*

Next we define multiplicity automata, which represents the function $f : \Sigma^* \to K$.

**Definition 3.2** *A multiplicity automaton $A$ of size $r$ defined with respect to a field $K$, an alphabet $\Sigma$, and an empty string $\lambda$ consists of:*

---
[1] Recall, a field is an algebraic structure in which the operations of addition, subtraction, multiplication, and division (except division by the zero element) are permitted, and the associative, commutative, and distributive rules hold.

- *a set of matrices $\{\mu_\sigma : \sigma \in \Sigma\}$, each of size $r \times r$ where $\mu_\sigma(i,j) \in K$.*
- *an $r$ tuple $\gamma = (\gamma_1, \ldots, \gamma_r) \in K^r$.*
- *a mapping $\mu$ which maps strings in $\Sigma^*$ to $r \times r$ matrices. More precisely, $\mu(\lambda) = I$ (where $I$ is the identity matrix), and, for $w \in \Sigma^*$, $\mu(w) = \mu_{\sigma_1} \cdot \mu_{\sigma_2} \cdots \mu_{\sigma_n}$.*
- *a function $f_A(w) = [\mu(w)]_1 \cdot \gamma$, where $[\mu(w)]_i$ is the ith row of $\mu(w)$.*

We now provide a concrete example of a deterministic automaton and show its multiplicity automaton representation.

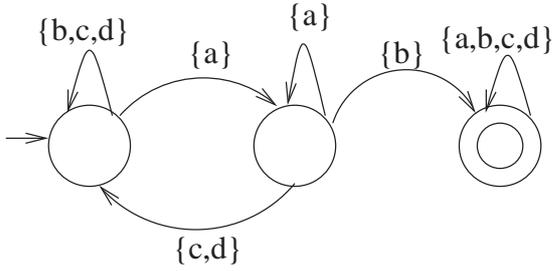

Figure 1: A deterministic automaton over $\Sigma = \{a, b, c, d\}$ that accepts a string only if it contains the substring $ab$, the accepting state is the rightmost state.

The multiplicity automaton of Figure 1 is of size 3 and its matrices are:

$$\mu_a = \begin{pmatrix} 0 & 1 & 0 \\ 0 & 1 & 0 \\ 0 & 0 & 1 \end{pmatrix}, \mu_b = \begin{pmatrix} 1 & 0 & 0 \\ 0 & 0 & 1 \\ 0 & 0 & 1 \end{pmatrix}$$

$$\mu_c = \begin{pmatrix} 1 & 0 & 0 \\ 1 & 0 & 0 \\ 0 & 0 & 1 \end{pmatrix}, \mu_c = \begin{pmatrix} 1 & 0 & 0 \\ 1 & 0 & 0 \\ 0 & 0 & 1 \end{pmatrix}$$

Set $\gamma = (0, 0, 1)$, since the rightmost state is the only accepting state.

We extend the previous example and show that probabilistic automata in general are a special case of multiplicity automata. Let the field be the reals and define $[\mu_\sigma]_{i,j}$ as the probability of moving from state $q_i$ to state $q_j$ with input $\sigma$, thus for a word $w = \sigma_1, \ldots, \sigma_n$, we have $\mu_w = \mu_{\sigma_1} \cdots \mu_{\sigma_n}$ and $[\mu_w]_{i,j}$ is the probability of moving from state $q_i$ to state $q_j$ with input string $w$. Let $\gamma$ will be vector of 1 in accepting states and 0 otherwise. Assuming that initial state is $s_1$, then $[\mu_w]_1 \cdot \gamma$ is the probability that $w$ is accepted by the probabilistic automaton.

We now introduce a theorem which relates the size of minimal automaton representing $f$ to the rank of $F$, its associate matrix. Recall the rank of a matrix is the number of linearly independent rows (or, equivalently, columns) in the matrix.

**Theorem 3.1** *(Carlyle and Paz [1971], Fliess [1974]) Let $f : \Sigma^* \to K$ such that $f \not\equiv 0$ and let F be the corresponding Hankel matrix. Then, the size $r$ of the smallest automaton A such that $f_A \equiv f$ satisfies $r = rank(F)$ (over the field K)*

## 4 Multiplicity Automata and Partially Observable MDP

In this section, we connect POMDPs and multiplicity automata. We show that POMDPs are actually MA. The connection is done through using the Predictive State Representation (PSR), which has received much attention in recent years (See for example Littman et al. [2001], Singh et al. [2003, 2004]).

The PSR of the POMDP describes the POMDP based on the possible futures rather than based on its structure. A future in a POMDP is a set of action observation pairs. A *test* is defined as a sequence of actions and observations $a_1, o_1, \ldots, a_k, o_k$, for every history[2] $h$ the probability of test being successful is defined as $\mathbf{P}(t|h) = \mathbf{P}(o_k, \ldots o_1|a_k, \ldots a_1, h)$. It is clear that if we enumerated all possible tests for all possible initial histories, then the model would be fully specified. However, the size of the representation described so far is infinite. Littman et al. [2001] showed that this representation can be done by a square matrix at size at most $|S|$. We will reprove this claim using the inherent connection between the PSR and the multiplicity automata.

**Theorem 4.1** *A POMDP is a special case of multiplicity automata.*

**Proof:** We will show how to construct a multiplicity automaton as was defined in definition 3.2 from a POMDP. The size of the automaton will be $|S|$, and $\Sigma$ is $A \times O \times R$. We let $\gamma$ be the initial distribution over the states, and for every $a \in A$ and $o \in O \times R$ we define a matrix $\mu_{a,o}$, where $\mu_{a,o}(i,j) = P(s_j, o|s_i, a)$. Recall that $\mu_\lambda = I$ and $\mu_{a_1,o_1,\ldots a_k,o_k} = \mu_{a_1,o_1} \cdots \mu_{a_k,o_k}$. We prove by induction that for, $[\mu_{a_1,o_1,\ldots a_k,o_k}]_i \cdot \gamma = \mathbf{P}(o_k, \ldots, o_1|a_k \ldots, a_1, s_i)$, where $[M]_i$ is the $i$th row and $\gamma = (1, 1, 1 \ldots, 1)$. First we prove the induction base case,

$$[\mu_{a_1,o_1}]_i \cdot \gamma = \sum_{j=1}^{n} \mathbf{P}(s_j, o_1|a_1, s_i) = \mathbf{P}(o_1|a_1, s_i)$$

Now assume the induction hypothesis holds for $k-1$ and prove for $k$

$$[\mu_{a_1,o_1,\ldots,a_k,o_k}]_i \cdot \gamma = [\mu_{a_1,o_1} \mu_{a_2,o_2,\ldots,a_k,o_k}]_i \cdot \gamma$$
$$= \sum_{j=1}^{n} [\mu_{a_1,o_1}]_{i,j} [\mu_{a_2,o_2,\ldots,a_k,o_k}]_j \cdot \gamma$$

---

[2] We will allow for history of size 1 to be an initial states

$$= \sum_{j=1}^{n} [\mu_{a_1,o_1}]_{i,j} \mathbf{P}(o_k, \ldots, o_2 | a_k, \ldots, a_2, s_j)$$

$$= \sum_{j=1}^{n} \mathbf{P}(s_j, o_1 | a_1, s_i) \mathbf{P}(o_k, \ldots, o_2 | a_k, \ldots, a_2, s_j)$$

$$= \sum_{j=1}^{n} \mathbf{P}(o_k, \ldots, o_1, s_j | a_k, \ldots, a_1, s_i)$$

$$= \mathbf{P}(o_k, \ldots, o_1 | a_k, \ldots, a_1, s_i)$$

$\square$

**Corollary 4.2** *POMDP can be represented as multiplicity automata of size $r$, $r \leq |S|$.*

**Proof:** By Theorem 4.1, we know that a POMDP is multiplicity automata by Theorem 3.1 its size is equal to the Hankel matrix rank. The rank of the Hankel matrix is bounded by $|S|$ since we can first write $|S|$ rows corresponding to all states, and these rows will span any other possible initial history. $\square$

An important note here is that the Hankel matrix corresponding to the multiplicity automata is essentially the same matrix used in the PSR literature. Thus the inherent connection between the approaches. In the next section, similarly to the PSR works we show how to construct a minimal submatrix of the Hankel matrix with the Hankel matrix rank. We denote the minimal basis as the set of rows and columns which consists the minimal submatrix of the Hankel matrix.

## 5 A Minimal Basis

While in the previous section we pointed out that there exists a minimal basis in size at most $|S|$ of both histories and tests, in this section we show how it can be found efficiently. We also show how to find efficiently a basis with additional properties — namely, a basis set that only needs "small" weights to span all states. This "small" weight property allows us to construct our planning algorithm using such a basis.

### 5.1 POMDP to Minimal Matrix Representation

In this subsection, we construct a minimal size matrix corresponding to a given POMDP.

Starting with notation. Given a set of tests $T = \{t_1, \ldots, t_m\}$ we define for every initial belief state $x$, a vector $F_x^T$ of size $m$, where $F_x^T(i) = \mathbf{P}(t_i | x)$, which denotes the probability of test $t_i$ starting from belief state $x$. We also define $\mathbf{P}(\lambda | x) = 0$ for any belief state $x$ (recall $\lambda$ is the empty string). For a state $s \in S$, we slightly abuse notation by writing $s$ to also mean the belief state which places all probability mass on the state $s$. So we write $F_s^T$, where

---

**Input**: POMDP=$(S, A, O, R, P)$
**Output**: Basis $B = \{b_1, \ldots, b_r\}$ and test basis
    $T = \{t_1, \ldots, t_r\}$
Initialize $B \leftarrow \{s_1\}, T \leftarrow \{\lambda\}$;
**while** *Exists $\sigma \in A \times O \times R$ and $b \in S \setminus B$, such that*
1. $F_b^T = \sum_{b_i \in B} \alpha_i F_{b_i}^T$
2. $\exists y \in T : P(\sigma \circ y | b) \neq \sum_{b_i \in B} \alpha_i P(\sigma \circ y | b_i)$
**do**
    $B = B \cup \{b\}$;
    $T = T \cup \{\sigma \circ y\}$;
**end**

**Algorithm 1**: Construction a minimal matrix representing POMDP

the initial belief state has all probability mass on state $s$. For a set of belief states $X$ and a set of Tests $T$, we denote $F_X^T$ as the set $\{F_x^T | x \in X\}$. Also note that the POMDP is represented by infinite matrix, but the rank of the matrix is finite and the algorithm finds a matrix with the minimal rank.

Before proving that the algorithm constructs the minimal basis, let us provide intuition behind the algorithm. First, this algorithm resembles the one for learning MA from equivalence and member queries given in Beimel et al. [2000] and the algorithm of Littman et al. [2001] where the predictive state representation was introduced. The algorithm begins by taking the test basis to be the empty test and for the initial history to be an arbitrary state. After phase $k$ the algorithm has $k$ histories and $k$ tests which are linearly independent [3] and should decide whether they are a basis and if not it should find an additional test and history and add them to the basis. We show that if we did not find a basis yet then an additional history can be found in $S$ and an additional test can be found as a one-step, extension test of the current basis tests. Thus, a basis can be computed efficiently.

**Lemma 5.1** *Algorithm 1 builds a submatrix of the Hankel matrix with rank equal to the Hankel matrix rank.*

**Proof:** We first show that at any stage $B$ is linearly independent with respect to the tests set $T$. We prove by induction, the basis holds trivially. Assume that the induction hypothesis holds for $B_\ell = \{b_1, \ldots, b_\ell\}$ and $T_\ell = \{t_1, \ldots, t_\ell\}$ and prove for $\ell + 1$. Since $F_{B_\ell}^{T_\ell}$ are linearly independent then $F_{B_\ell}^{T_{\ell+1}}$ are linearly independent as well. By the way we choose $t_{\ell+1}$ and $b_{\ell+1}$ we are assured $F_{b_{\ell+1}}^{T_{\ell+1}}$ is linearly independent of $F_{B_\ell}^{T_{\ell+1}}$. Now we have to prove that our matrix is not smaller than the minimal matrix. By the construction of Beimel et al. [2000] we know that that if the matrix rank is less than the Hankel matrix rank, then there

---

[3] This implies that the $\alpha$s calculated by the algorithm are unique

exists a counterexample $w$ and a test $\sigma \circ y$, $y \in T$ such that

- $F_w^T = \sum_{b_i \in B} \alpha_i F_{b_i}^T$
- $P(\sigma \circ y | w) \neq \sum_{b_i \in B} \alpha_i P(\sigma \circ y | b_i)$

Since every $w$ is a linear combination of $S = \{s_1, \ldots, s_n\}$, then $F_w^T$ can be represented as linear combination $F_S^T$. Thus the rank of $s_1, \ldots, s_n$ is larger than the rank of $B$ with respect to $T \cup \{\sigma \circ y\}$, and we can find a state $s_i$ which is linearly independent of $B$ with respect to $T$ thus adding $s_i$ to $B$ increases its dimension. $\square$

**Lemma 5.2** *Algorithm 1 terminates in*
$O(r^4 |S|^2 |A||O||R|)$ *steps.*

**Proof:** Since we build a basis of size at most $|S|$, there at most $|S|$ iterations. In each iteration we check for every combination of state, action and observation if they satisfy condition 1 and 2 which takes $O(r^3)$ $\square$

### 5.2 A Modified Belief MDP

Similarly to the *belief MDP*, we would like to define the *modified belief MDP*.

**Definition 5.1** *Let $r$ be the size of a basis $B = (b_1, \ldots, b_r)$ and let $T = \{t_1, \ldots, t_r\}$ be the test basis. The modified belief MDP states are vector of size $r$ such that $x = (\alpha_1, \ldots, \alpha_r)$ if and only if $F_x^T = \sum_{i=1}^r \alpha_i F_{b_i}^T$. Let $x = (\alpha_1, \ldots, \alpha_r)$ and $y = (\beta_1, \ldots, \beta_r)$ two modified belief states, then $\mathbf{P}(x, a, y) = \sum_{b' | F_{b'}^T = \sum_{i=1}^r \beta_i F_{b_i}^T} \mathbf{P}(b, a, b')$ for any $b = \sum_{i=1}^r \alpha_i F_{b_i}^T$. Let $x = (\alpha_1, \ldots, \alpha_r)$ and $b$ a any belief state such that $F_b^T = \sum_{i=1}^r \alpha_i F_{b_i}^T$, then the reward of action $u$ in state $x$ is $\sum_{r' \in R} \mathbf{P}(r|u, b) r'$.*

The optimal policy in the *modified belief MDP* as the optimal policy in the *belief MDP* is the optimal policy in the original POMDP, since in both cases the representation is exact. Note that although each state in the *modified belief MDP* is represented by a coefficients vector, this representation differs from the standard belief space representation as the coefficients of the basis states $F_s^T = \sum_{s_i \in B} \beta_i F_{s_i}^T$ can be very large and negative.

Therefore, building a grid with respect to the *modified belief MDP* will depend on the largest coefficient and on the matrix rank (which is at most $|S|$). In the next section we would show how to improve our basis and overcome the dependence on the largest coefficient.

### 5.3 Barycenteric Spanner - Improving the basis

In this subsection we introduce how to transform the basis outputted by Alg. 1 to a "better" basis by using Awerbuch and Kleinberg [2004] technique. More formally, the main theorem of the subsection shows that we can find a basis such that for any *belief state* largest coefficient is bounded by 2.

**Theorem 5.3** *Let $r$ be the rank of the multiplicity automata representing the POMDP and let $T$ be the set of tests outputted by Alg. 1. There exists a set of states $B \subset S$, where $|S| = r$, such that, for any belief state $b$, the vector $F_b^T$ can be uniquely represented as $F_b^T = \sum_{\tilde{b} \in B} \alpha_i F_{\tilde{b}}^T$, with $\max_i |\alpha_i| \leq 2$.*

We first provide a definition of barycenteric spanner.

**Definition 5.2 (Awerbuch and Kleinberg [2004])** *A $C$-barycenteric spanner is a basis $B$, which spans a convex set $S$, such that every $x \in S$ can be represented as $x = \sum_{b_i \in B} \alpha_i b_i$, where $|\alpha_i| \leq C$.*

Awerbuch and Kleinberg [2004] showed that if $S$ is a compact set that contained in $\mathbb{R}^d$ but is not contained in linear subspace of smaller dimension then there exists a 1-barycenteric spanner. If we could transform that basis outputted by Alg. 1 to a barycenteric spanner, we can the discretize the *modified belief MDP* with dependence only on the automaton rank since the largest coefficient is constant. The next algorithm shows how to construct efficiently an "almost" spanner as was done in Awerbuch and Kleinberg [2004]. We denote the matrix of basis initial histories $B$ and Tests $T$ as $M(B, T)$, where $M(B, T)_{i,j} = \mathbf{P}(t_j | b_i)$.

---

**Input** : POMDP=$(S, A, O, R, P)$
    Basis $B = \{b_1, \ldots, b_r\}$
    and test basis $T = \{t_1, \ldots, t_r\}$
**Output**: Basis $\tilde{B} = \{\tilde{b}_1, \ldots, \tilde{b}_r\}$
**Initialize:** $\tilde{b}_i = b_i, \forall i$;
**while** $\exists x \in S \setminus \tilde{B}, i \in \{1, \ldots, r\}$ *satisfying*
$det(M(x \cup B \setminus \tilde{b}_i, T)) > 2 det(M(\tilde{B}, T))$ **do**
    $\tilde{b}_i = x$;
**end**

**Algorithm 2**: Construction a 2-Barycenteric spanner

---

Note the spanner spans the the set $\{F_s^T : s \in S\}$, which is identical to span the set $\{F_b^T : b \text{ is a belief state}\}$.

This next lemma proves that the Algorithm builds a 2-Barycenteric spanner and its proof can found in Awerbuch and Kleinberg [2004]. For intuition, note that this algorithm must terminate since there exists a 1-barycenteric spanner, which achieves the largest determinant, and in each iteration the determinant grows by at least a factor of two.

**Lemma 5.4** *Alg. 2 constructs a 2-Barycenteric spanner for the set $\{F_s^T : s \in S\}$.*

Next we show that Algorithm 2 terminates in polynomial time.

**Lemma 5.5** *Algorithm 2 terminates in $O(|S|r^4 \log r)$ steps.*

**Proof:** By Lemma 2.5 in Awerbuch and Kleinberg [2004] we know that the loop repeats $O(r \log r)$. In each iteration we calculate the determinant of a square matrix of size $r$ at most $S$ times, thus each iteration takes at most $|S|r^3$ time steps. □

Now we are ready to prove the main theorem of this subsection, which demonstrates the advantage of the basis produced by Alg. 2 over the basis produced by Alg. 1 in which $\alpha_i$ is unbounded.

**Proof of Theorem 5.3:**

Let $b = \sum_{i=1}^{|S|} \beta_i s_i$ any belief state.

$$\begin{aligned} F_b^T &= \sum_{i=1}^{|S|} \beta_i F_{s_i}^T = \sum_{i=1}^{|S|} \beta_i \sum_{j=1}^{r} \gamma_j(i) F_{\tilde{b}_j}^T \\ &= \sum_{j=1}^{r} \sum_{i=1}^{|S|} \beta_i \gamma_j(i) F_{\tilde{b}_j}^T = \sum_{j=1}^{r} \alpha_j F_{\tilde{b}_j}^T \end{aligned}$$

Now we observer that

$$\alpha_j = \sum_{j=1}^{r} \sum_{i=1}^{|S|} \beta_i \gamma_j(i) \leq \max_i \gamma_j(i) \leq 2,$$

where the first inequality comes from the fact that $\sum_{i=1}^{|S|} \beta_i = 1$ and that $\beta_i > 0$ for every $i$, and the second inequality follows from the fact that $s_i$s were used in building the barycenteric spanner, and thus for every $i, j$ $|\gamma_j(i)| \leq 2$. □

## 6 Planning Algorithm

We are finally ready to describe our planning algorithm, Algorithm 3. The planning algorithm uses the first two algorithms to construct a basis of size $r$ of initial states and $r$ tests. With each *modified belief state* fully represented by a linear combination of the basis states with coefficient bounded in $[-2, 2]$, note that in contrast to the standard belief state the coefficient can be negative. Instead of discretizing the *Belief MDP* we discretize the *modified belief MDP*, which is built with respect to the basis (see definition 5.1). Our planning algorithm will then discretize it will compute the optimal policy for this MDP.

### 6.1 The Algorithm

**Input**: POMDP $P = (S, A, O, R, P)$

**Output**: An $\epsilon$-optimal policy $\pi$

Run Alg. 1 on $(P)$ and obtain $T = \{t_1, \ldots t_r\}$ and $\tilde{B} = \{b_1, \ldots, b_r\}$ ;

Run Alg. 2 on $(P, \tilde{B}, T)$ and obtain $B = \{b_1, \ldots, b_r\}$;

Build a grid based MDP, $\tilde{M}$ with respect to the basis $B = \{b_1, \ldots, b_r\}$ and tests $T = \{t_1, \ldots, t_r\}$ ;

Each state in the MDP is a vector of size $r$, $(\alpha_1, \ldots, \alpha_r)$, where $\alpha_i = m\tilde{\epsilon}, m \in \{-2/\tilde{\epsilon}, \ldots, 0, \ldots, 2/\tilde{\epsilon}\}, \tilde{\epsilon} = \epsilon/r$ ;

Compute an optimal policy $\pi$ for $\tilde{M}$.

**Algorithm 3**: Planning

Now we state our main theorem.

**Theorem 6.1** *Given a POMDP, Algorithm 3 computes an $\frac{\epsilon}{(1-\gamma)^4}$-optimal policy in time polynomial in $|S|, |A|, |O|, |R|, \epsilon$ and exponentially only in $r$, where $r$ is the size of the multiplicity automaton representing the POMDP.*

### 6.2 Algorithm Analysis

We first show that the algorithm is efficient, i.e., polynomial in $|S|, |A|, |O|$ and exponential in $r$ the matrix rank. Later we provide performance guarantees for the the algorithm, i.e., that it produces an $\epsilon$-optimal policy.

**Lemma 6.2** *Algorithm 3 terminates in $O\left(r^4 |S|^2 |A| |O| |R| \log r + \left(\frac{2r}{\epsilon}\right)^{3r}\right)$ steps.*

**Proof:** Let's first analyze the time. By Lemmas 5.2 and 5.5 running Alg. 1 and Alg 2 takes $O(r^4|S|^2|A||O||R|\log r)$. The grid MDP contains $\left(\frac{2r}{\epsilon}\right)^r$ states and thus we can solve it in $\left(\frac{2r}{\epsilon}\right)^{3r}$. □

Similarly to Subsection 2.2 we would like to show that if two states are mapped to the same grid state then they are close to each other in the sense that the optimal value is close between them.

**Lemma 6.3** *Let $b$ and $b'$ two belief states that are mapped to the same state in the grid MDP, then for every action $u \in A$ $|Q^*(b, u) - Q^*(b', u)| \leq \frac{\epsilon}{1-\gamma}$*

**Proof:** We show here that if $b$ and $b'$ are mapped to the same state in the $\epsilon$-discretized modified belief MDP then for every policy $\pi^4$, $|V^\pi(b) - V^\pi(b')| \leq \epsilon$. Let $b$ be a belief state such that $F_b^T = \sum_{i=1}^{r} \alpha_i F_{b_i}^T$ and $b'$ such that $F_{b'}^T = \sum_{i=1}^{r} \beta_i F_{b_i}^T$, then since $b$ and $b'$ are in the same state in the discretized MDP $\|\alpha - \beta\|_1 \leq \epsilon$. Let $R^t$ be the return

---
[4]Note that $\pi$ is not necessarily stationary

in test $t$, then we can rewrite

$$\begin{aligned} V^\pi(b) &= \sum_t \mathbf{P}(t|b,\pi)R^t = \sum_t \sum_{i=1}^r \alpha_i \mathbf{P}(t|b_i,\pi)R^t \\ &\leq \sum_t \sum_{i=1}^r \beta_i \mathbf{P}(t|b_i,\pi)R^t + \frac{\epsilon}{1-\gamma} \\ &= V^\pi(b') + \frac{\epsilon}{1-\gamma} \end{aligned}$$

Similarly we can show that $V^\pi(b) \geq V^\pi(b') - \frac{\epsilon}{1-\gamma}$. □

Now similarly to the proof of Theorem 2.1 we apply Lemmas 2.3 and 2.4 to derive the following lemma.

**Lemma 6.4** *Algorithm 3 computes an $\frac{\epsilon}{(1-\gamma)^4}$-optimal policy.*

Now the proof of Theorem 6.1 follows from Lemmas 6.2 and 6.4.

### Acknowledgements

This work was supported in part by the IST Programme of the European Community, under the PASCAL Network of Excellence, IST-2002-506778, and by a grant from the Israel Science Foundation and an IBM faculty award. This publication only reflects the authors' views.